\UseRawInputEncoding

\documentclass{article}

\usepackage{PRIMEarxiv}
\usepackage[utf8]{inputenc} 
\usepackage[T1]{fontenc}    
\usepackage{hyperref}       
\usepackage{url}            
\usepackage{booktabs}       
\usepackage{amsfonts}       
\usepackage{nicefrac}       
\usepackage{microtype}      
\usepackage{lipsum}
\usepackage{fancyhdr}       
\usepackage{graphicx}       
\graphicspath{{media/}}     

\usepackage{multirow}
\usepackage{amsmath}
\usepackage[ruled]{algorithm2e}
\usepackage{bm}
\usepackage{float}  
\usepackage{subfigure}  

\title{KDSM: An uplift modeling framework based on knowledge distillation and sample matching
}

\author{
    Chang Sun\textsuperscript{1}, Qianying Li\textsuperscript{2}, Guanxiang Wang\textsuperscript{1} Sihao Xu\textsuperscript{2}, Yitong Liu\textsuperscript{*1}\\
    1:Beijing University of Posts and Telecommunications\\
    2:Beijing Qunar Software Technology Co., Ltd.\\
    sc1998@bupt.edu.cn, qianying.li@qunar.com, WangGX@bupt.edu.cn \\
    sihao.xu@qunar.com, liuyitong@bupt.edu.cn
}

\begin{document}
\maketitle

\begin{abstract}
Uplift modeling aims to estimate the treatment effect on individuals, widely applied in the e-commerce platform to target persuadable customers and maximize the return of marketing activities. Among the existing uplift modeling methods, tree-based methods are adept at fitting increment and generalization, while neural-network-based models excel at predicting absolute value and precision, and these advantages have not been fully explored and combined. Also, the lack of counterfactual sample pairs is the root challenge in uplift modeling. In this paper, we proposed an uplift modeling framework based on Knowledge Distillation and Sample Matching (KDSM). The teacher model is the uplift decision tree (UpliftDT), whose structure is exploited to construct counterfactual sample pairs, and the pairwise incremental prediction is treated as another objective for the student model. Under the idea of multitask learning, the student model can achieve better performance on generalization and even surpass the teacher. Extensive offline experiments validate the universality of different combinations of teachers and student models and the superiority of KDSM measured against the baselines. In online A/B testing, the cost of each incremental room night is reduced by 6.5\%. 
\end{abstract}

\section{Introduction}
\begin{figure}[b]
\centerline{\includegraphics[width=5cm]{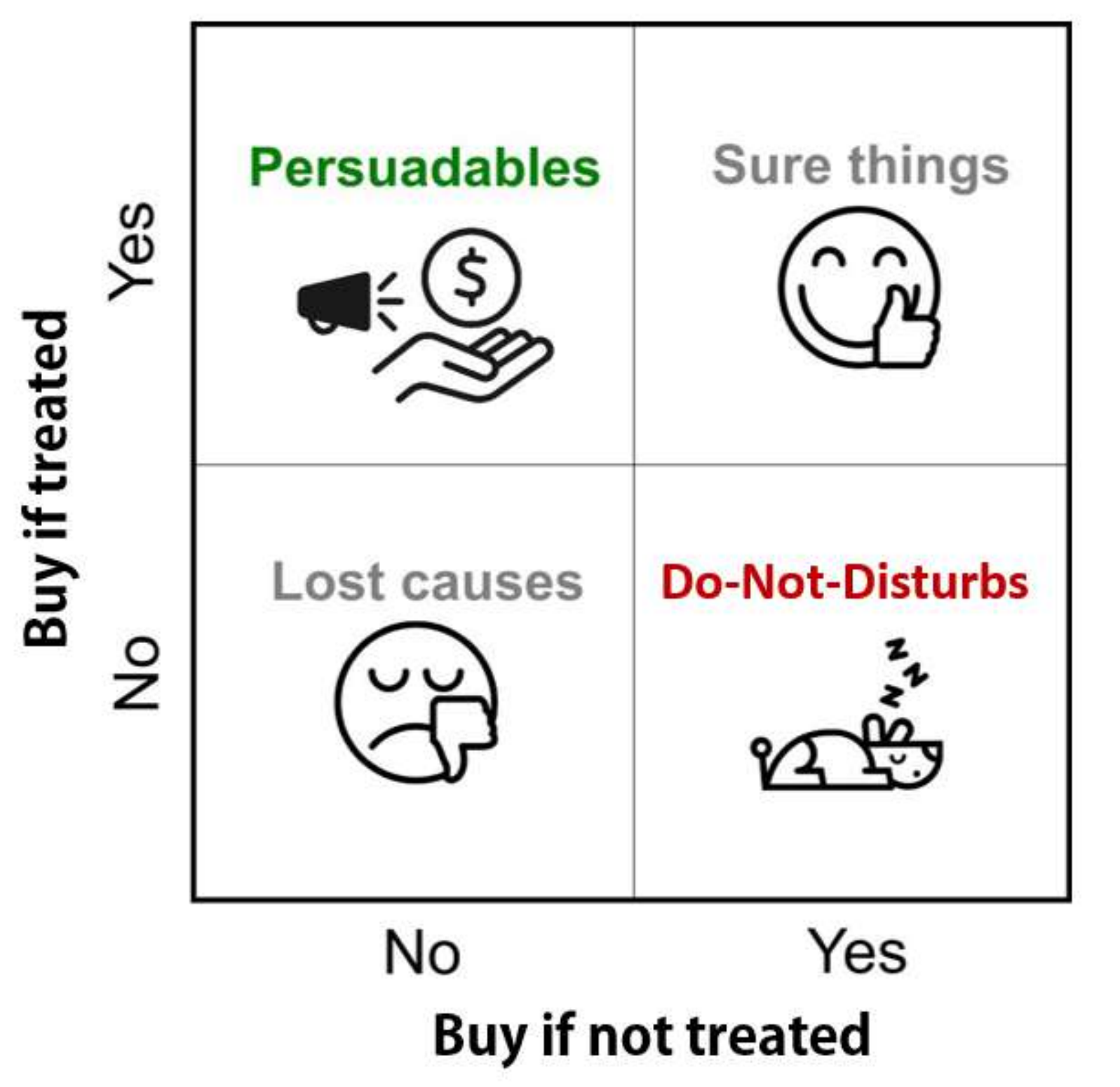}}
\caption{The four categories of customers \cite{devriendt2018literature}.  }
\label{sec:fig0}
\end{figure}

Uplift modeling aims to estimate the incremental impact of a treatment on an individual outcome, and the increment is also known as \emph{individual treatment effect} (ITE) or \emph{uplift}. Since uplift modeling can be conducive to targeting sub-groups of people and providing personalized strategies, it is widely applied in marketing \cite{radcliffe2007using}, social science \cite{imai2013estimating, kunzel2019metalearners} and medicine treatment \cite{zhang2017mining}. Specifically, e-commerce companies attempt to predict the uplift of promotion campaigns to focus on the \emph{persuadable} customers and avoid \emph{Do-Not-Disturb} customers \cite{devriendt2018literature}, maximizing the return of investment of a campaign, as shown in Figure \ref{sec:fig0}.

Uplift models can be divided into two categories according to their objects to model: response model and incremental model. A response model estimates the probability of a customer to respond \cite{devriendt2018literature}, which takes \emph{treatment} as a feature and has to predict twice to obtain the incremental impact \cite{lo2002true}, while an incremental model predicts the increment directly \cite{rzepakowski2010decision}. In industrial practice, response models are usually realized by neural networks, for they can take both continuous and discrete features and automatically learn representations and feature interactions in the hidden space. However, due to the sparsity of data (take click or purchase as an example), deep neural networks tend to be overfitting with unsatisfactory generalization \cite{zhou2018deep}. On the contrary, a prevailing series of incremental models are decision trees with modified splitting criteria and pruning techniques \cite{devriendt2018literature}. The performance of tree-based models highly relies on the hyper-parameters: deep decision trees suffer from overfitting, while shallow trees deal with fewer features with better generalization and worse precision \cite{quinlan1986induction}.

Inspired by the characteristics of response models and incremental models, we propose a universal uplift modeling framework based on knowledge distillation and sample matching (KDSM). In our approach, the teacher model is an UpliftDT \cite{rzepakowski2010decision}, while the student model is a neural network, thus the knowledge of increment is transferred into the student model. We conduct the proposed KDSM framework on both open source data and real hotel reservation business data, and the experimental results confirm the effectiveness. Online A/B testing proved that KDSM can improve the room night increments and reduce the cost of promotion. The main contributions of this paper are as follows:

\begin{enumerate}
    
    \item We utilize the idea of knowledge distillation \cite{hinton2015distilling} and multitask learning \cite{caruana1997multitask}, combining the sensitivity of NN-based response models and the generalization of tree-based incremental models so that the student model can be superior to the teacher model.
    
    \item We propose a counterfactual sample matching method relying on the structure of UpliftDT, which can be treated as a plug-and-play data augment method. 

    \item We implement extensive experiments on two offline datasets and online A/B testing. Experimental results demonstrate that KDSM can outperform the baselines on offline data and reduce the cost of sending coupons in real business. 
    
\end{enumerate}

The structure of this paper is as follows: Section 2 introduces the related work about uplift modeling and knowledge distillation; Section 3 formulates the problem; Section 4 states the details of KDSM framework; Section 5 shows the experimental results, which prove the superiority and universality of our proposed method, and Section 6 declares our conclusion.

\section{Related Work}

\subsection{Uplift modeling}

The main task of uplift modeling is to estimate the causal effect of treatment on individuals in different sub-populations. Uplift modeling methods can be divided into two major categories according to their origins: one is to extend existing machine learning algorithms, and the other is to design methods specifically for uplift modeling estimation \cite{zhang2021unified}.

Meta-learner is a commonly used method in uplift modeling, which decomposes estimating the Conditional Average Treatment Effect (CATE) into several sub-regression problems that can be solved by one or more base-learners. Each base learner can be implemented with existing supervised machine learning algorithms, such as logistic regression, support vector machine \cite{cortes1995support}, etc. \emph{S-learner} \cite{lo2002true,athey2015machine} (S for single) means taking the concatenation of treatment and covariates as the input features and outcome variable as the target to train a single response model. \emph{T-learner} \cite{radcliffe2011real,nassif2013uplift,hansotia2002incremental} (T for two) makes an improvement by training two models on the treatment and control group separately. Künzel et al. \cite{kunzel2019metalearners} and Nie et al. \cite{2020Quasi} proposed X-learner and R-learner respectively based on S-learner and T-learner. In addition, the class transformation method creates a new dummy variable $Y^*$ from the observed variable $Y$, thereby transforming uplift into the conditional expectation of the transformed outcome $Y^*$ \cite{zhang2021unified,athey2015machine}.

In tailored methods for uplift modeling, tree-based methods are the most prevailing. One of the main advantages of the tree models is their strong interpretability \cite{zhang2021unified}, and the key to building an uplift tree is designing an appropriate splitting criterion. As one of the earliest tree-based models, Uplift Incremental Value Modelling \cite{hansotia2002incremental} has a splitting criterion aiming to maximize the difference of CATE between the left and right child nodes. Rzepakowski and Jaroszewicz \cite{rzepakowski2010decision} proposed three splitting criteria based on KL divergence, $\chi^2$ divergence, and Euclidean distance (ED) in UpliftDT. However, the main drawbacks of tree-based methods are: I) The construction process is greedy and does not return an optimal tree. Therefore, one tree and an alternative tree (by slightly perturbing data) from the same algorithm can differ significantly \cite{zhang2021unified}. II) Ensemble methods based on bagging \cite{breiman1996bagging} and random forests \cite{breiman2001random}, such as uplift bagging \cite{soltys2015ensemble,guelman2012random}, uplift random forest \cite{guelman2015uplift}, and causal forests \cite{wager2018estimation} can alleviate the first point, but they extend the training and inference time and lose the interpretability. III) The performance of the tree models depends on the hyper-parameters to a great extent.

For better locating the framework proposed in this paper, note that KDSM distills the knowledge from a tree-based model to a response model of S-Learner. Even though there are two models involved, the two models are not trained respectively for the treatment and control group. Therefore, the framework is an extension of S-Learner methods rather than T-Learner.

\subsection{Knowledge distillation}

Knowledge distillation (KD) methods \cite{hinton2015distilling} have been widely applied in various fields of machine learning since the idea was proposed in 2015. Hinton et al. use KD to encourage the outputs of the small distilled model to approximate that of the large model \cite{hinton2015distilling}. Similarly, the authors of LwF \cite{li2017learning} perform KD to learn new tasks while keeping knowledge of old tasks. In the field of recommendation and marketing, in 2020, Yichao Wang et al. \cite{wang2020practical} proposed distillation learning to solve the catastrophic forgetting problem in the incremental learning of CTR prediction models. In 2022, Kang et al. \cite{kang2022personalized} proposed a method to distill the preference knowledge in a balanced way without relying on any assumption on the representation space.

In summary, the idea of knowledge distillation is suitable for transferring implicit "knowledge" from one model (teacher) to another model (student). In this way, the student model can obtain new information from the teacher model on the premise of retaining the original learning ability. KD combines the respective advantages of the two models, which can be an effective approach for lightweight deployment \cite{hinton2015distilling}, task migration \cite{li2017learning}, and incremental learning \cite{wang2020practical}.

\section{Problem definition}

Uplift modeling can be defined under the potential outcome framework \cite{rubin1974estimating}. $T \in \{0,1\}$ denotes a binary treatment variable. In marketing, the treatment can be a campaign or a coupon. 

\begin{equation}
T_i = 
\begin{cases}
1, \text{if $i$-th subject receives the treatment;}\\
0, \text{otherwise}
\end{cases}
\end{equation}

Let binary variable $Y_i$ denote the outcome of $i$-th subject, which means whether a customer would purchase a product. Every subject has two potential outcomes: $Y_i(0)$ stands for the potential outcome if $T_i = 0$, and $Y_i(1)$ stands for the potential outcome if $T_i=1$. For subject $i$, the ITE of treatment $T$ is defined as the difference between the two potential outcomes: 

\begin{equation}
    \tau_i = Y_i(1) - Y_i(0)
\end{equation}

Unfortunately, only one of the two potential outcomes can be observed, therefore ITE is not identifiable. Conditional average treatment effect (CATE) is the average treatment effect conditioning on a set of covariates that describe the subjects, which is the best estimator of ITE. Specifically, let $X = \boldsymbol{x} \in \mathbb{R}^k$ be a covariate vector describing the characteristics of subjects. The CATE is defined as:

\begin{equation}
    \tau(\boldsymbol{x}) = \mathbb{E}[Y(1)-Y(0)|X = \boldsymbol{x}]
\end{equation}

In randomized controlled trials (RCTs), uplift is transformed to the difference of conditional expectations, which can be estimated unbiasedly.

\begin{equation}
    Uplift(\boldsymbol{x})=\mathbb{E}(Y|T=1,X=\boldsymbol{x})- \mathbb{E}(Y|T=0,X=\boldsymbol{x})
    \label{eq4}
\end{equation}

\section{Methods}

In this section, we introduce the teacher model and the student model respectively, then we elaborate the knowledge distillation framework and the sample matching strategy. At last, a theoretical analysis of the effectiveness is provided.

\begin{figure}[bt]
\centerline{\includegraphics[width=\linewidth]{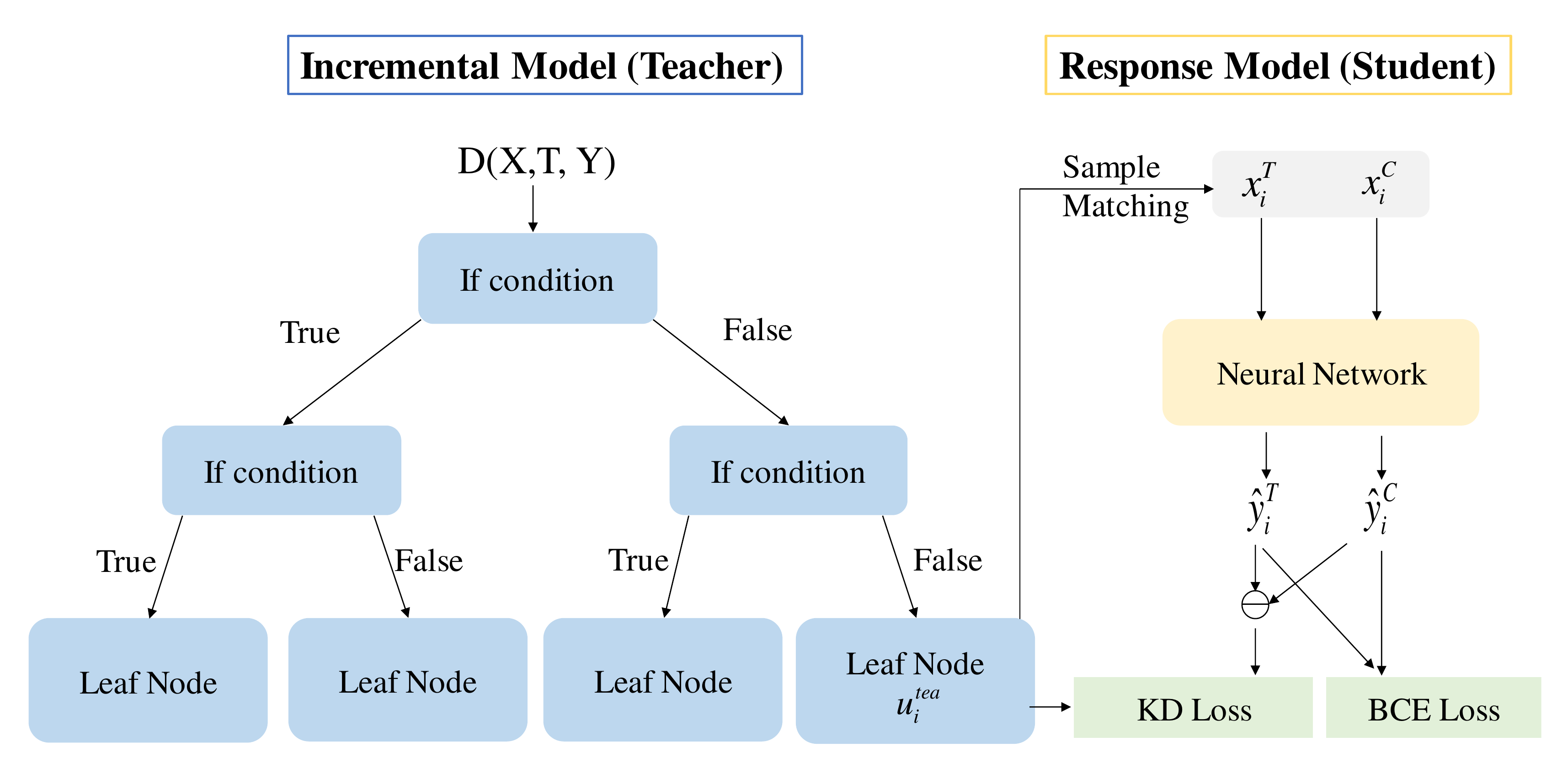}}
\caption{The structure of KDSM framework.}
\label{sec:fig1}
\end{figure}

\subsection{Teacher model}

The teacher model is the UpliftDT \cite{rzepakowski2010decision}, the most prevailing approach in incremental models, which directly models and predicts the increment. While training, the tree model recursively splits the node into two child nodes by a splitting criterion. During prediction, subjects fall into different leaf nodes and get their uplift value, as shown on the left of Figure \ref{sec:fig1}.

The construction of the tree model in this framework can be flexible: there is no limitation in the splitting criteria. UpliftDT can be fitted by maximizing ED, KL divergence, etc. ED is superior because it is more stable and has the important property of being symmetric \cite{rzepakowski2010decision}. The splitting criteria of ED can be expressed as:

\begin{equation}
C^{Eu} = \frac{n_L}{n}\hat{\tau}_L^2 + \frac{n_R}{n}\hat{\tau}_R^2
\end{equation}

\noindent where $n$ stands for the number of subjects in a node, $n_L$ and $n_R$ denote the number of subjects in its left and right child nodes, and $\hat{\tau}_L$ and $\hat{\tau}_R$ are estimated as the within-node CATEs in the training data.

As for the depth of trees, a shallow depth may result in a single prediction for a great portion of samples, while a decision tree with deep depth tends to overfit the training set, resulting in poor generalization \cite{quinlan1986induction}. Therefore, the depth of trees is set to 5 after our experiments.

\subsection{Student model}

The student model is a single response model implemented by a neural network, denoted as $M_s$. The input is the concatenated eigenvector $[X, T]$, and $Y$ is the target. The label of the $i$th sample $(x_i,t_i)$ is $y_i$, and the model estimates the value of the conditional probability $P(Y|T=t,X=x)$:

\begin{equation}
    \hat{y}_i=M_s(x_i,t_i)
\end{equation}

When Y is binary, the loss function of the response model is binary cross-entropy (BCE) loss. The BCE loss should be expressed as Equation \ref{eq5}.

\begin{equation}
    \mathcal{L}_{BCE}(y_i,\hat{y}_i)= {{y_i} \times \ln } {\hat y}_i - (1 - {y_i}) \times \ln (1 - {\hat y_i})
    \label{eq5}
\end{equation}

To predict uplift, the student model predicts the response of $T=1$ and $T=0$ respectively, and uplift is their difference, as shown in Equation \ref{eq6}.

\begin{equation}
   \hat{Uplift}\left(x\right)=M_S\left(X=x,T=1\right)-M_S\left(X=x,T=0\right)
   \label{eq6}
\end{equation}

\subsection{Knowledge distillation framework}

In the KDSM framework, we train an UpliftDT first, and it splits training samples into different leaf nodes. Then, we implement sample matching within leaf nodes, and the $i$th pair of counterfactual samples includes one sample from the treatment group and one sample from the control group, denoted as $x_i^T$ and $x_i^C$. The specific matching method is illustrated in Section 4.4. The prediction given by the student model for these two samples is $\hat{y}_i^T$ and $\hat{y}_i^C$, thus the uplift, $u_i^{stu}$ can be expressed as:

\begin{equation}
\begin{aligned}
   u_i^{stu} &= M_S(X=x_i^T,T=1)-M_S(X=x_i^C,T=0) = \hat{y}_i^T-\hat{y}_i^C
   \label{eq7}
\end{aligned}
\end{equation}

Uplift of the $i$th pair predicted by the teacher model $M_T$ is $u_i^{tea}$, and the pairwise KD loss is designed to be the L2 loss between $u_i^{tea}$ and $u_i^{stu}$.

\begin{equation}
   \mathcal{L}_{KD}(\hat{y}_i^T, \hat{y}_i^C) =  (u_i^{tea}-u_i^{stu})^2
   \label{eq8}
\end{equation}

Thus the total pairwise loss function can be expressed as Equation \ref{eq9}.

\begin{equation}
\mathcal{L}(\hat{y}_i^T, \hat{y}_i^C) = \mathcal{L}_{BCE}( y_i^T,\hat{y}_i^T)+ \mathcal{L}_{BCE}( y_i^C,\hat{ y}_i^C)+\lambda \mathcal{L}_{KD}(\hat{y}_i^T, \hat{y}_i^C)
\label{eq9}
\end{equation}

The first two terms are the hard loss, learning to fit the response. The third term is the soft loss, learning the the knowledge of incremental "taught" by the teacher model. KDSM transfers the knowledge of increment to the response model so that the neural network can combine the information of increment with its own ability to fit the response. The generalization of the student model is improved, while the advantages of precision and sensitivity remain.

\begin{algorithm}[h]
\caption{Framework of KDSM.}
\KwIn{Training set $D(X,T,Y)$, a pre-trained teacher model $M_T$, KD weight $\lambda$.}
\KwOut{Trained student model $M_S$.}
Randomly initialize student model $M_s$, whose parameters are denoted as $\theta_S$. UpliftDT $M_T$ splits samples in $D$ into some subsets, which is corresponding to the leaf nodes. \\
\While{not convergence}{
Set an empty dataset denoted as $D^\prime$.\\
\For{subset $s \in D$}{
Randomly match sample pair ($y_i^T$,$y_i^C$) without replacement and put them into dataset $D^\prime$.
}
    \ForEach{batch $B \in D^\prime$}{
      \ForEach{sample pair $(y_i^T, y_i^C)$}{
        Compute $\hat{y}_i^T=M_S(x_i^T,T=1)$, $\hat{y}_i^C=M_S(x_i^C,T=0)$.\\
        Compute $\mathcal{L}(\hat{y}_i^T, \hat{y}_i^C)=\mathcal{L}_{BCE}(y_i^T,\hat{y}_i^T)+\mathcal{L}_{BCE}(y_i^C,\hat{y}_i^C)+\lambda \mathcal{L}_{KD}(\hat{y}_i^T, \hat{y}_i^C)$.
      }
  Update student model $\theta_S \gets \frac{\partial \mathcal{L}}{\partial \theta_S}$.
    }
}
\end{algorithm}

\subsection{Sample matching}

Why is sample matching necessary for implementing knowledge distillation? An optional method is to distill based on single samples, that is to predict $M_S(X=x_i,T=1)-M_S(X=x_i,T=0)$ for a single sample $x_i$ (no matter whether it is treated or not), and encourage it to be close to $u_i^{tea}$. However, there is no label for the counterfactual sample, whose BCE loss can not be computed. Thus the answer is positive, and both the counterfactual samples should be real.

We propose sample matching within leaf nodes: randomly selecting a sample from the treatment and the control group, respectively, and considering them as a counterfactual sample pair. It makes pairwise KD loss available, meanwhile the BCE loss of the two samples can be computed respectively. It is easy and intuitive: these two samples fall into the same leaf node because they meet the same conditions on the most distinguishing features (the features involved in the path from the root node to the leaf node). Therefore, we can regard them as approximate counterfactual samples.

There are different methods to achieve sample matching, and random matching is one of the easiest to achieve. To ensure each sample is sent into the model equiprobably, eliminate the random deviation and increase the diversity of sample pairs, we conduct random matching before each epoch. In this way, plenty of counterfactual sample pairs can be produced without changing the total amount of sample, which can be regarded as an effective data enhancement method.

\section{Experiments}

\subsection{Datasets}

\textbf{Criteo:} Criteo published CRITEO-UPLIFT dataset v2 \cite{diemert2021large}, with 12 numerical features and over 13 million samples. We focus on the treatment and conversion fields, and their specific business scenarios have been hidden. We sampled 1,000,000 data from this dataset, the number of samples of the treatment group and control group is 1:1, and the training set, test set, and validation set are divided into 3:1:1.

\textbf{Production:} Production dataset comes from Qunar Company's offline log of the hotel reservation business, which includes samples from the treatment group and the control group. The subjects in the treatment group receive a 9\% coupon, while the subjects in the control group do not receive any coupon. It was collected in 40 consecutive days, and there are over 560,000 samples in total. The number of samples from the treatment group and control group is 1:1, and the training set, test set, and validation set are divided into 3:1:1. We selected 50 features according to their correlation to the outcome variable, including 34 numerical features and 16 category features.

\textbf{Data Protection Statement: }(1) The data used in this study did not involve any personally identifiable information (PII). (2) The data used in this study are processed by data abstraction and encryption, and the researchers cannot recover the original data. (3) The data is only used for academic research, and sufficient data protection is carried out during the experiment to prevent data leakage.

\subsection{Evaluation metrics and experimental settings}

\textbf{Evaluation metrics.} The uplift curve and the Qini curve are usually used for the evaluation of uplift modeling when the outcome variable is binary \cite{zhang2021unified}. Let $\pi$ be a descending ordering of subjects according to their estimated treatment effects. We use $\pi(k)$ to denote the first $k$ subjects. Let $R_{\pi(k)}$ be the count of positive outcomes in $\pi(k)$ and let $R_{\pi(k)}^{T=1}$ and $R_{\pi(k)}^{T=0}$ be the number of positive outcomes in the treatment and control groups, respectively. Finally, let $N_{\pi(k)}^{T=1}$ and $N_{\pi(k)}^{T=0}$ be the number of subjects in the treatment and control groups from $\pi(k)$. The uplift curve and Qini curve can be expressed as:

\begin{equation}
    \text{uplift}(k)=(\frac{R_{\pi(k)}^{T=1}}{N_{\pi(k)}^{T=1}}-\frac{R_{\pi(k)}^{T=0}}{N_{\pi(k)}^{T=0}})\cdot (N_{\pi(k)}^{T=1}+N_{\pi(k)}^{T=0})
\end{equation}

\begin{equation}
    \text{Qini}(k)=R_{\pi(k)}^{T=1}-R_{\pi(k)}^{T=0}\frac{N_{\pi(k)}^{T=1}}{N_{\pi(k)}^{T=0}}
\end{equation}

The uplift curve and Qini curve can be drawn by varying $k$. Since the uplift and Qini curves are similar in terms of their shape \cite{gutierrez2017causal}, we will illustrate the Qini curve, AUUC (area under uplift curve), and Qini coefficient (area under Qini curve) as our evaluation metrics.

\textbf{Experimental settings.} The UpliftDT is realized by CausalML \cite{chen2020causalml}, while the neural networks are realized by PyTorch \cite{paszke2019pytorch}, and most of the baselines and their parameters are referenced from the benchmark published by Criteo \cite{diemert2021large}. The initial learning rate is 1e-2, and early stopping is deployed according to the AUUC on the validation set. On the Criteo dataset, $batchsize$ is 8192, and the patience of early stopping is 20 epochs. The factor of learning rate decay is 0.1, and the patience is 3 epochs. On the production dataset, $batchsize$ is 1024, and the patience of early stopping is 18 epochs.

\subsection{Performance across different teachers and students}

To verify the universality of our knowledge distillation framework, we implement experiments on two teacher models and four student models. For teacher models, we adopt the criterion of KL divergence and ED, and Table \ref{table0} shows their performance.

\begin{table}[hbp]\centering
\caption{The results of Uplift decision trees of different split criteria on Criteo and Production dataset.}
\label{table0}
\begin{tabular}{c|cc|cc}
\hline
\multirow{2}{*}{} & \multicolumn{2}{c|}{KL} & \multicolumn{2}{c}{ED} \\
 & AUUC & Qini & AUUC & Qini \\ \hline
Criteo & 0.7110 & 0.1292 & \textbf{0.8900} & \textbf{0.4180} \\ \hline
Production & 0.5485 & 0.0040 & \textbf{0.5825} & \textbf{0.0797} \\ \hline
\end{tabular}
\end{table}

For student models, since the logic of predicting the probability of purchasing is similar to that of CTR, we choose the following four classic CTR prediction models as student models:

\begin{itemize}

\item DCN (Deep \& Cross Network)\cite{RuoxiWang2017DeepC} was proposed in 2017, which explicitly applies feature crossing at each layer, requires no manual feature engineering, and adds negligible extra complexity to the DNN model.

\item AutoInt\cite{WeipingSong2019AutoIntAF} maps both the numerical and categorical features into the same low-dimensional space, then uses a multi-head self-attentive neural network with residual connections to explicitly model the feature interactions in the low-dimensional space.

\item DeepFM \cite{guo2017deepfm} was proposed in 2017 and it replace the "Wide" part in Wide \& Deep \cite{HengTzeCheng2016WideD} by FM \cite{SteffenRendle2010FactorizationM}. The FM module provides the capability to learn feature interactions and its deep neural network provides the representation ability.

\item xDeepFM (eXtreme Deep Factorization Machine) \cite{lian2018xdeepfm} was proposed in 2018, and the authors designed the Compressed Interaction Network (CIN), which aims to generate feature interactions explicitly at the vector-wise level.

\end{itemize}

\begin{table}[hbp]\centering
\caption{The results of different teacher models across different student models.}
 \label{table1}
\begin{tabular}{ccccccc}
\hline
\multicolumn{7}{c}{Criteo} \\ \hline
\multicolumn{1}{c|}{Student} & \multicolumn{2}{c|}{w/o KD} & \multicolumn{2}{c|}{KD with KL} & \multicolumn{2}{c}{KD with ED} \\ \cline{2-7} 
\multicolumn{1}{c|}{Model} & \multicolumn{1}{c|}{AUUC} & \multicolumn{1}{c|}{Qini} & \multicolumn{1}{c|}{AUUC} & \multicolumn{1}{c|}{Qini} & \multicolumn{1}{c|}{AUUC} & Qini \\ \hline
\multicolumn{1}{c|}{DCN} & \multicolumn{1}{c|}{0.9223} & \multicolumn{1}{c|}{0.4641} & \multicolumn{1}{c|}{0.9361} & \multicolumn{1}{c|}{0.4764} & \multicolumn{1}{c|}{\textbf{0.9676}} & \textbf{0.5104} \\ \hline
\multicolumn{1}{c|}{AutoInt} & \multicolumn{1}{c|}{0.8913} & \multicolumn{1}{c|}{0.4323} & \multicolumn{1}{c|}{0.9167} & \multicolumn{1}{c|}{0.4579} & \multicolumn{1}{c|}{\textbf{0.9556}} & \textbf{0.5014} \\ \hline
\multicolumn{1}{c|}{DeepFM} & \multicolumn{1}{c|}{0.9087} & \multicolumn{1}{c|}{0.4496} & \multicolumn{1}{c|}{0.9134} & \multicolumn{1}{c|}{0.4540} & \multicolumn{1}{c|}{\textbf{0.9784}} & \textbf{0.5256} \\ \hline
\multicolumn{1}{c|}{xDeepFM} & \multicolumn{1}{c|}{0.9100} & \multicolumn{1}{c|}{0.4514} & \multicolumn{1}{c|}{0.9287} & \multicolumn{1}{c|}{0.4556} & \multicolumn{1}{c|}{\textbf{0.9287}} & \textbf{0.4723} \\ \hline
\end{tabular}

\begin{tabular}{ccccccc}
\hline
\multicolumn{7}{c}{Production} \\ \hline
\multicolumn{1}{c|}{Student} & \multicolumn{2}{c|}{w/o KD} & \multicolumn{2}{c|}{KD with KL} & \multicolumn{2}{c}{KD with ED} \\ \cline{2-7} 
\multicolumn{1}{c|}{Model} & \multicolumn{1}{c|}{AUUC} & \multicolumn{1}{c|}{Qini} & \multicolumn{1}{c|}{AUUC} & \multicolumn{1}{c|}{Qini} & \multicolumn{1}{c|}{AUUC} & Qini \\ \hline
\multicolumn{1}{c|}{DCN} & \multicolumn{1}{c|}{0.6473} & \multicolumn{1}{c|}{0.1580} & \multicolumn{1}{c|}{\textbf{0.6877}} & \multicolumn{1}{c|}{\textbf{0.1885}} & \multicolumn{1}{c|}{0.6761} & 0.1769 \\ \hline
\multicolumn{1}{c|}{AutoInt} & \multicolumn{1}{c|}{0.6599} & \multicolumn{1}{c|}{0.1715} & \multicolumn{1}{c|}{0.6751} & \multicolumn{1}{c|}{0.1786} & \multicolumn{1}{c|}{\textbf{0.6771}} & \textbf{0.1886} \\ \hline
\multicolumn{1}{c|}{DeepFM} & \multicolumn{1}{c|}{0.6145} & \multicolumn{1}{c|}{0.1273} & \multicolumn{1}{c|}{0.6292} & \multicolumn{1}{c|}{0.1356} & \multicolumn{1}{c|}{\textbf{0.6320}} & \textbf{0.1456} \\ \hline
\multicolumn{1}{c|}{xDeepFM} & \multicolumn{1}{c|}{0.6447} & \multicolumn{1}{c|}{0.1580} & \multicolumn{1}{c|}{0.6826} & \multicolumn{1}{c|}{0.1862} & \multicolumn{1}{c|}{\textbf{0.6771}} & \textbf{0.1890} \\ \hline
\end{tabular}
\end{table}

\begin{figure}[tbp]
    \centering  
    \subfigure[UpliftDT(ED)-DeepFM on Criteo dataset]{
    \includegraphics[width=0.48\linewidth]{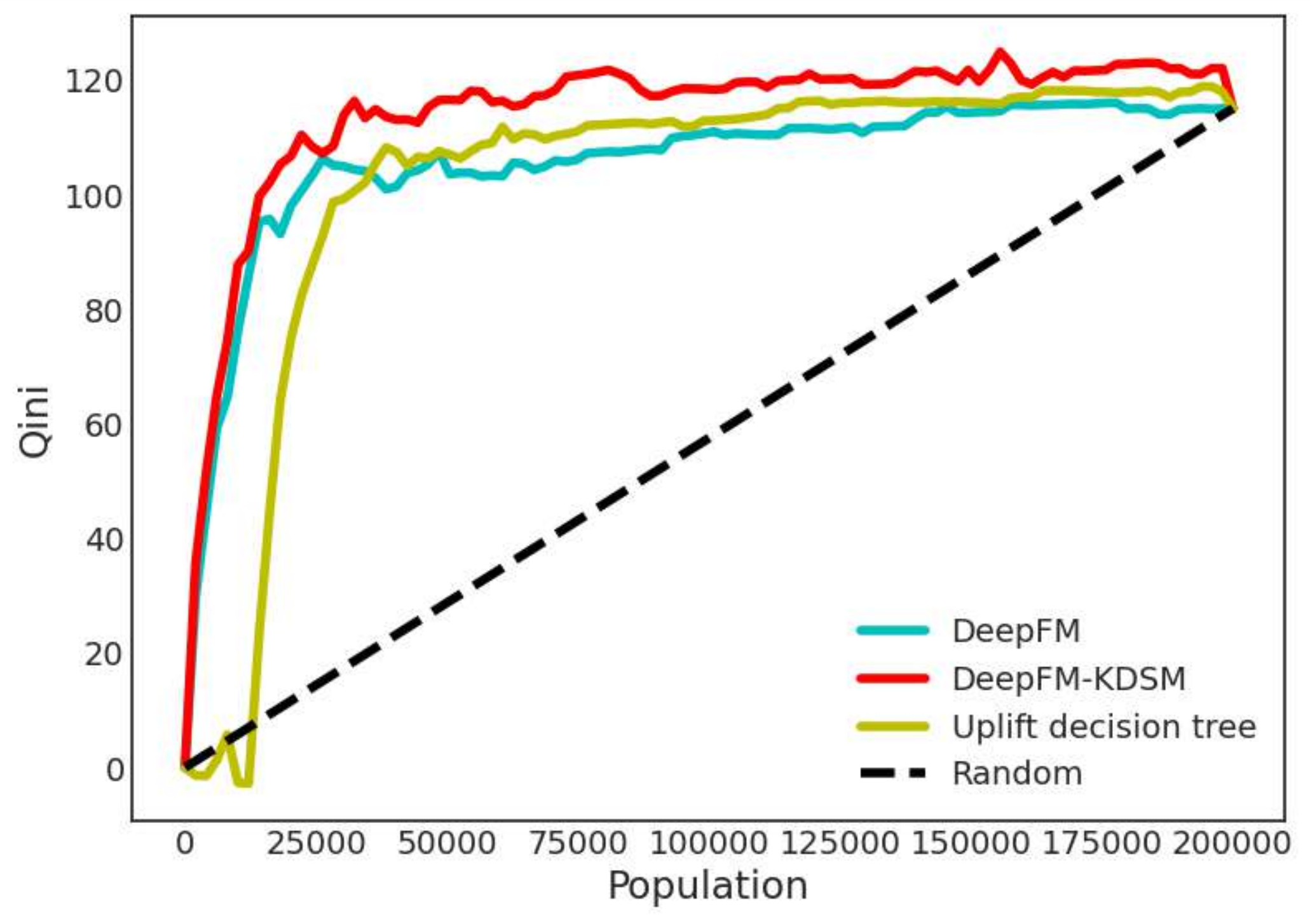}}
    \subfigure[UpliftDT(KL)-xDeepFM on Production dataset]{
    \includegraphics[width=0.48\linewidth]{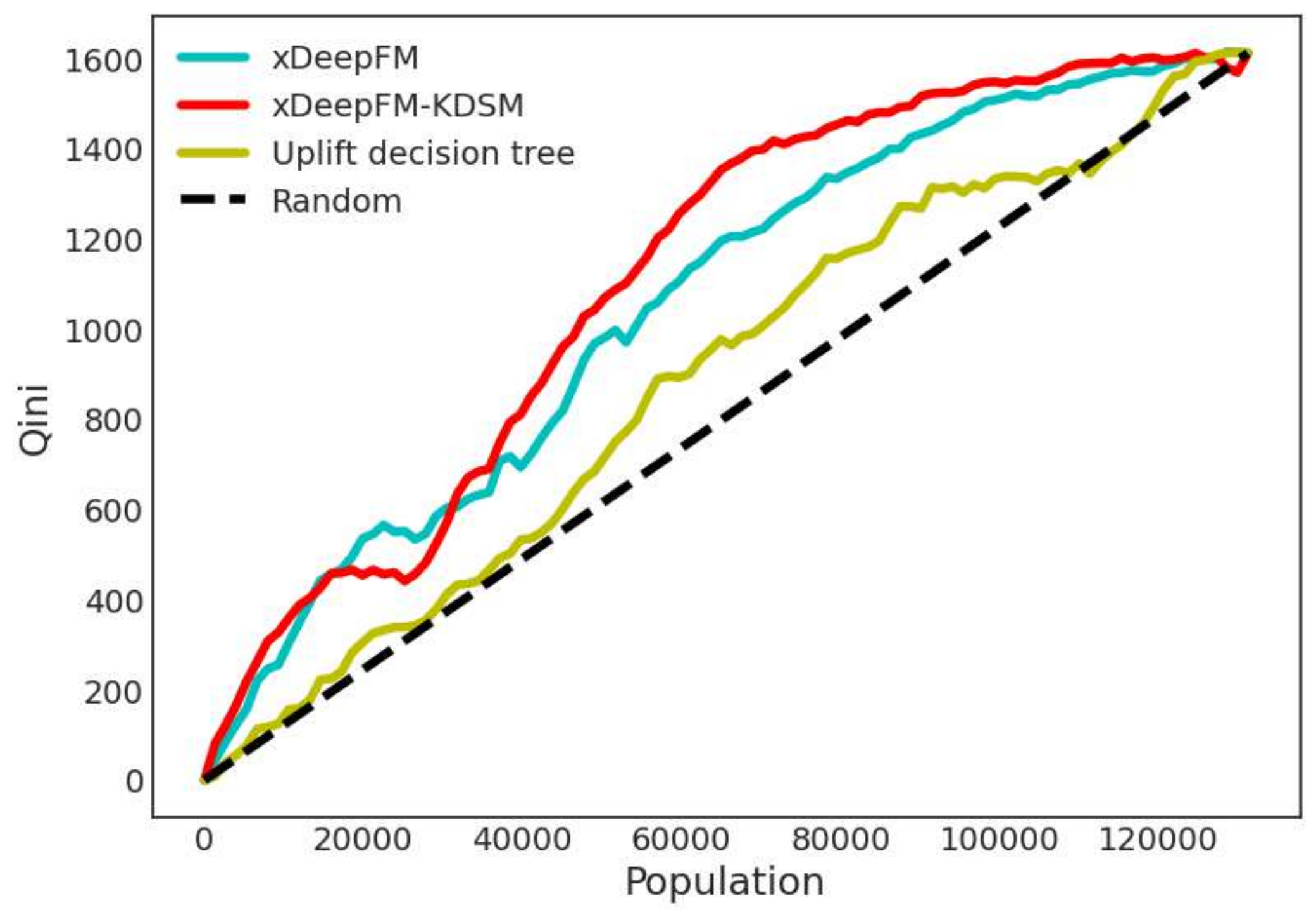}}
    \caption{The histogram of the uplift prediction.}
    \label{fig3}
\end{figure}

As shown in Table \ref{table0} and Table \ref{table1}, the knowledge generated by both the "KL tree" and the "ED tree" can bring gains to the performance of the student model on both the two datasets. The Qini coefficient increased by 16.9\% on the Criteo dataset and 19.6\% on Qunar Company's production dataset at most. The results are consistent with our intuitions: students can be improved to different extents after learning, and talented students would still be excellent in the class. On the other hand, for a single student, the more superior his teacher is, the more improvement he can get, so as a student model in knowledge distillation.

Due to the space limit, we only show the results of DeepFM-ED on the Criteo dataset in Figure \ref{fig3}(a) and xDeepFM-KL on the production dataset in Figure \ref{fig3}(b) as we have similar observations on other combinations of teachers and students.

In Figure \ref{fig3}(a), the Qini curve can be divided into two segments: rapid increasing segment and plateau. Rapid increasing corresponds to the persuadable customers ($uplift > 0$), and the plateau corresponds to the \emph{sure things} and the \emph{lost causes} ($uplift \approx 0$). Note that subjects are ranked descendingly according to their uplift prediction. Compared with DeepFM, DeepFM-KDSM can recognize more persuadable customers (ranking about 10,000th-25,000th). UpliftDT performs a negative effect on the top-ranking customers, which tends to be because some leaf nodes are wrongly over-fitted.

In Figure \ref{fig3}(b), XDeepFM-KDSM performs similarly to xDeepFM for the top 40,000 users but represents its superiority over the latter users. As can be seen, KL-UpliftDT performs terribly and only works for the middle-ranking users, but it still provides knowledge of the increment and improves the performance of xDeepFM on the middle-ranking users.

Furthermore, it can be observed that the Criteo dataset is much easier to be fitted, and it is difficult for models to learn from the features and distinguish the persuadable customers from the production data. Despite the complexity of real-world data, the KDSM framework can improve the performance of the student model.

\subsection{Hyper-parameter sensitivity}

In knowledge distillation frameworks, the most significant hyper-parameter is the weight of KD loss, $\lambda$. To find the optimal $\lambda$, we implement Bayesian optimization \cite{pelikan1999boa} to locate a rough range and carry out experiments on some fixed values. The results are shown in Table \ref{table2}. The second row stands for the weight of the KD loss, and the "0" columns stand for using the student model without KD loss.

\begin{table*}[htbp]\centering
\caption{The results of hyper-parameter sensitivity on $\lambda$.}
\label{table2}
\begin{tabular}{c|c|ccc|cccc}
\hline
Split & Student & \multicolumn{3}{c|}{Criteo} & \multicolumn{4}{c}{Production} \\ \cline{3-9}
Criterion & Model & 0 & 100 & 1000 & 0 & 0.1 & 0.5 & 1 \\ \hline
\multirow{4}{*}{ED} & DCN & 0.9223 & 0.9424 & \textbf{0.9676} & 0.6473 & \textbf{0.6761} & 0.6462 & 0.6542 \\
 & AutoInt & 0.8913 & 0.9286 & \textbf{0.9556} & 0.6599 & 0.6763 & \textbf{0.6771} & 0.6671 \\
 & DeepFM & 0.9087 & 0.9336 & \textbf{0.9784} & 0.6145 & \textbf{0.6320} & 0.6318 & 0.6195 \\
 & xDeepFM & 0.9100 & 0.9127 & \textbf{0.9287} & 0.6447 & 0.6659 & \textbf{0.6771} & 0.6657 \\ \hline
\multirow{4}{*}{KL} & DCN & 0.9223 & 0.9424 & \textbf{0.9676} & 0.6473 & \textbf{0.6877} & 0.6604 & 0.6347 \\
 & AutoInt & 0.8913 & 0.9107 & \textbf{0.9167} & 0.6599 & 0.6748 & \textbf{0.6751} & 0.6578 \\
 & DeepFM & 0.9087 & 0.9102 & \textbf{0.9134} & 0.6145 & \textbf{0.6292} & 0.6271 & 0.6102 \\
 & xDeepFM & 0.9100 & 0.9127 & \textbf{0.9287} & 0.6447 & 0.6773 & \textbf{0.6826} & 0.6573 \\ \hline
\end{tabular}
\end{table*} 

As is shown in Table \ref{table2}, different $\lambda$ can bring gain to the student models to different extents in most cases. There is no universally optimal value, and the optimal $\lambda$ should be searched and adjusted according to the specific network structure and dataset. Generally, the order of magnitude of $\lambda \times L_{KD}$ should be kept same as the hard loss.

\subsection{The impact of the weight of the KD loss}

To explore the influence of KD on the distribution of uplift prediction, we take the KL tree as the teacher, xDeepFM as the student, and the Production dataset as examples to observe the distribution of their output with histograms.

\begin{figure}[tbp]
	\centering  
	\subfigure[KL UpliftDT]{
		\includegraphics[width=0.45\linewidth]{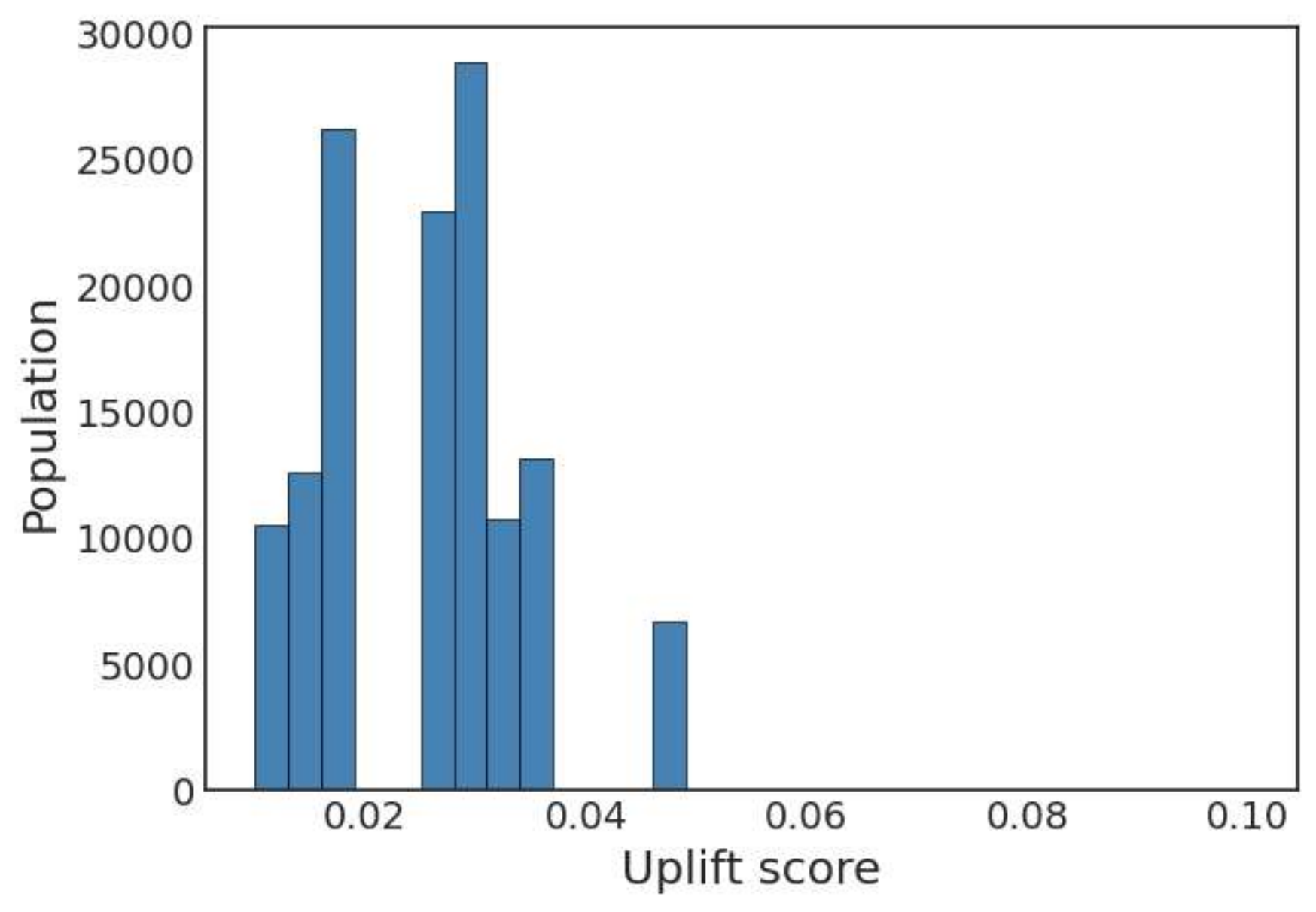}}
	\subfigure[xDeepFM]{
		\includegraphics[width=0.45\linewidth]{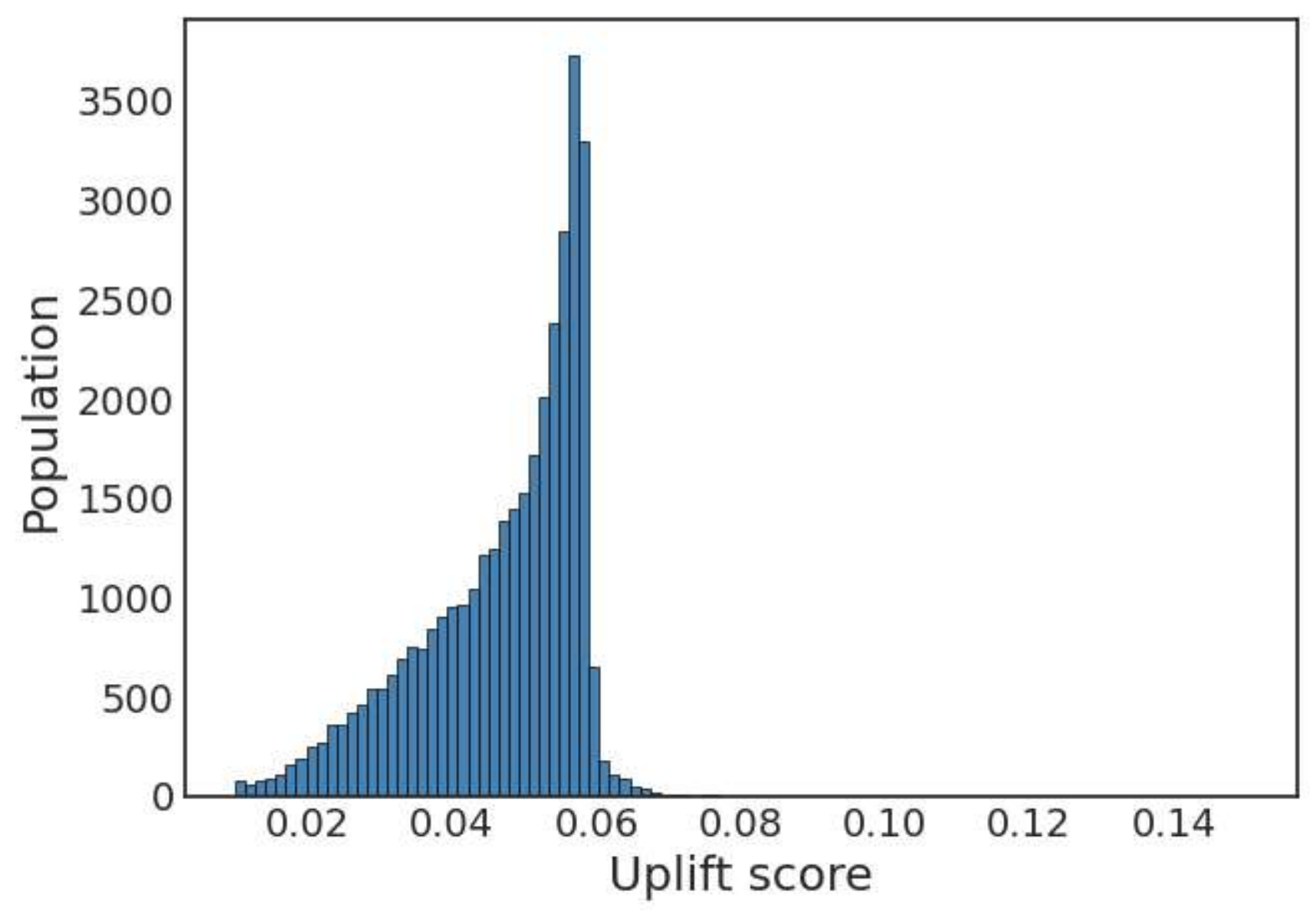}}
	\\
	\subfigure[xDeepFM-KDSM, $\lambda=0.1$]{
		\includegraphics[width=0.45\linewidth]{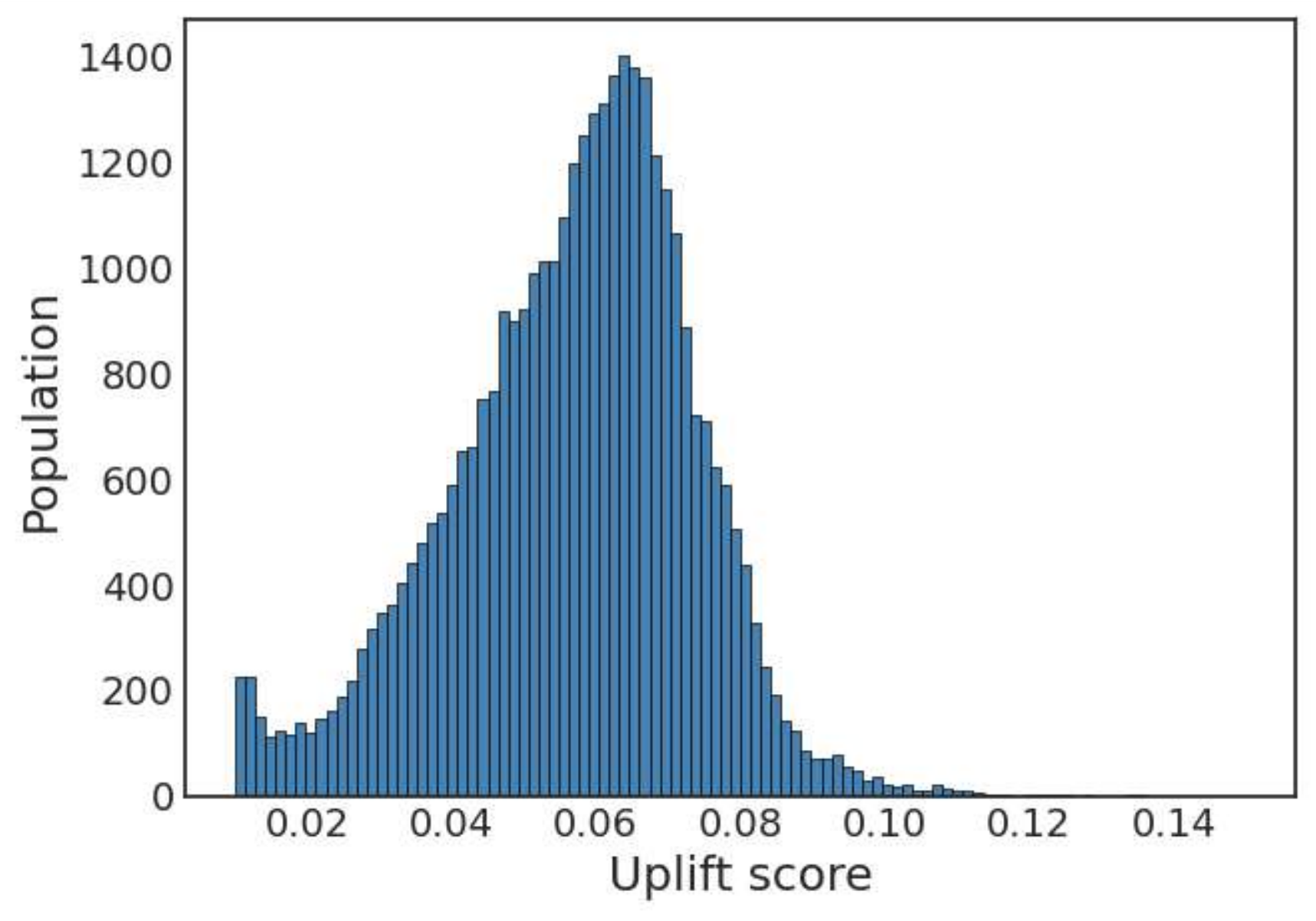}}
	\subfigure[xDeepFM-KDSM, $\lambda=0.5$]{
		\includegraphics[width=0.45\linewidth]{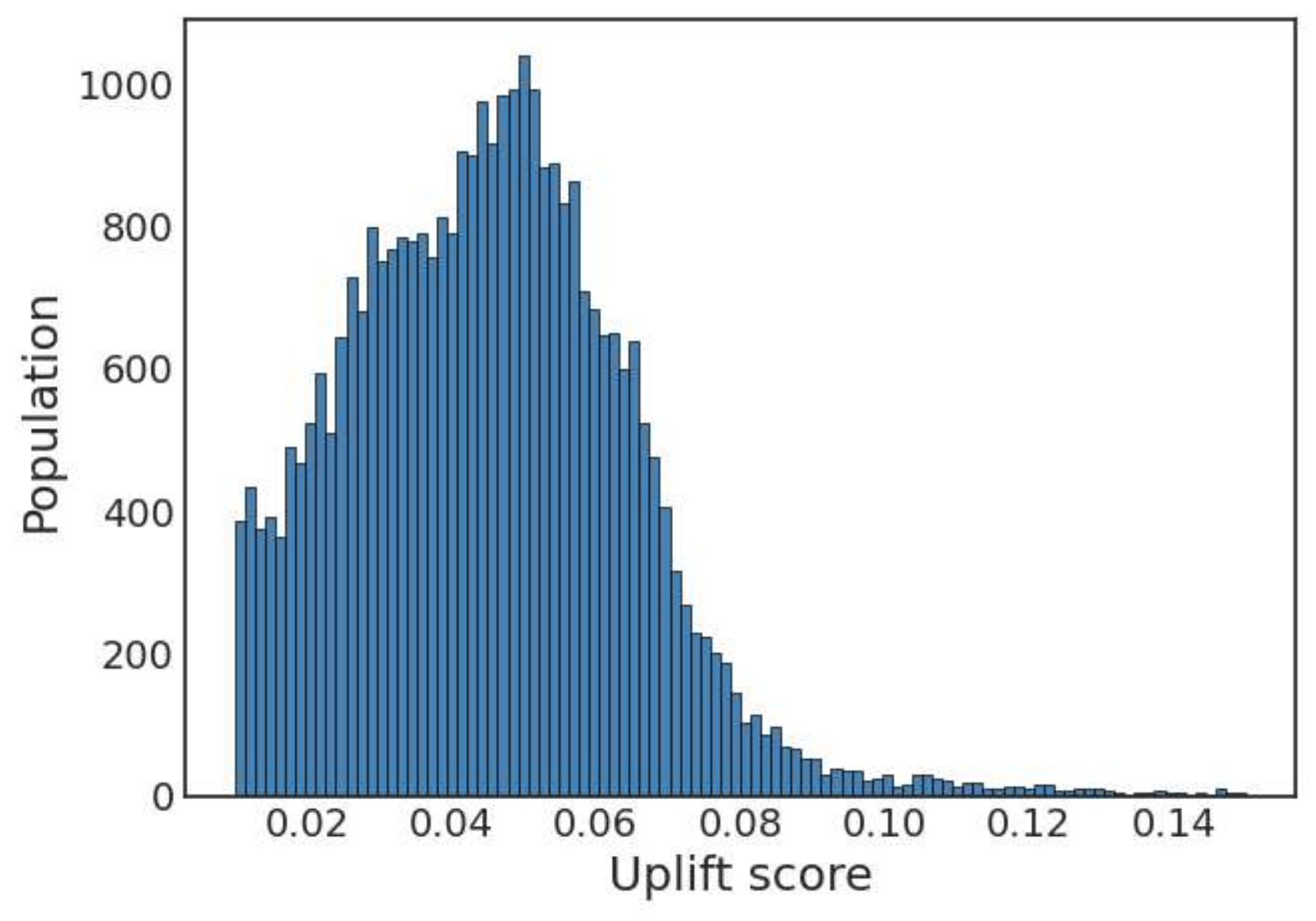}}
	\caption{The histogram of the uplift prediction.}
	\label{fig4}
\end{figure}

Figure \ref{fig4} is the histogram of uplift prediction. Compared with the distribution of the prediction of student model w/o KD, the distributions in (c) and (d) are pulled closer to the teacher model by the KD loss. Also, when the weight $\lambda$ increases, the left shift of the KDSM output distribution becomes obvious.

\subsection{Ablation study on knowledge distillation strategy}

To verify the necessity of sample matching, we carried out three experiments: (a) Student model without knowledge distillation. (b) Knowledge distillation on single samples, as introduced in the first paragraph of Section 4.4, denoted as KDSS. (c) Matching the counterfactual sample pairs randomly before each epoch, denoted as KDSM. The UpliftDT splits by the criterion of ED. The weight of soft loss is set to 1000 on the Criteo dataset, and 0.5 on the Production dataset.

\begin{table}[tbp]\centering
\caption{The comparison of different knowledge distillation strategies.}
\label{table3}
\begin{tabular}{c|l|cc|cc}
\hline
Student & \multicolumn{1}{c|}{KD} & \multicolumn{2}{c|}{Criteo} & \multicolumn{2}{c}{Production} \\ \cline{3-6} 
Model & \multicolumn{1}{c|}{Strategy} & AUUC & Qini & AUUC & Qini \\ \hline
\multirow{3}{*}{DCN} & (a)w/o KD & 0.9223 & 0.4641 & 0.6473 & 0.1500 \\
 & (b)KDSS & 0.9266 & 0.4679 & 0.6456 & 0.1482 \\
 & (c)KDSM & \textbf{0.9676} & \textbf{0.5104} & \textbf{0.6604} & \textbf{0.1604} \\ \hline
\multirow{3}{*}{AutoInt} & (a)w/o KD & 0.8913 & 0.4323 & 0.6599 & 0.1715 \\
 & (b)KDSS & 0.9149 & 0.4563 & 0.6743 & 0.1761 \\
 & (c)KDSM & \textbf{0.9556} & \textbf{0.5014} & \textbf{0.6771} & \textbf{0.1886} \\ \hline
\multirow{3}{*}{DeepFM} & (a)w/o KD & 0.9087 & 0.4496 & 0.6145 & 0.1273 \\
 & (b)KDSS & 0.9180 & 0.4591 & 0.6115 & 0.1145 \\
 & (c)KDSM & \textbf{0.9784} & \textbf{0.5256} & \textbf{0.6318} & \textbf{0.1442} \\ \hline
\multirow{3}{*}{xDeepFM} & (a)w/o KD & 0.9100 & 0.4514 & 0.6447 & 0.1580 \\
 & (b)KDSS & 0.9228 & 0.4641 & 0.6398 & 0.1431 \\
 & (c)KDSM & \textbf{0.9287} & \textbf{0.4723} & \textbf{0.6771} & \textbf{0.1890} \\ \hline
\end{tabular}
\end{table}

As can be observed in Table \ref{table3}, random matching before each epoch has the best performance on both the Criteo dataset and the Production dataset. "KDSS" is futile and even causes negative effects because the student model can be confused by the fake labels of the counterfactual samples.

\subsection{Comparison to other methods}

To investigate the effectiveness of our framework KDSM, we compare our method with several baselines:

\begin{itemize}
 \item \textbf{TM} (Two-Model) \cite{hansotia2002incremental}, a framework using two separate probabilistic models to fit the outcomes of control and treatment groups.
 
 \item \textbf{MOM} (Modified Outcome Method) \cite{athey2015machine}, which is also called as transformed outcome.
 
 \item \textbf{SDR} (Shared Data Representation) \cite{betlei2018uplift}, a multi-task framework considering predicting outcomes in control and treatment groups as the related tasks.
 
 \item \textbf{X-Learner} \cite{kunzel2019metalearners}, an outstanding method among meta-learners, which is efficient when the numbers of units are imbalanced in different groups.

\end{itemize}

\begin{figure}[tbp]
    \centering  
    \subfigure[The Qini curves of different methods on the Criteo dataset.]{
    \includegraphics[width=0.48\linewidth]{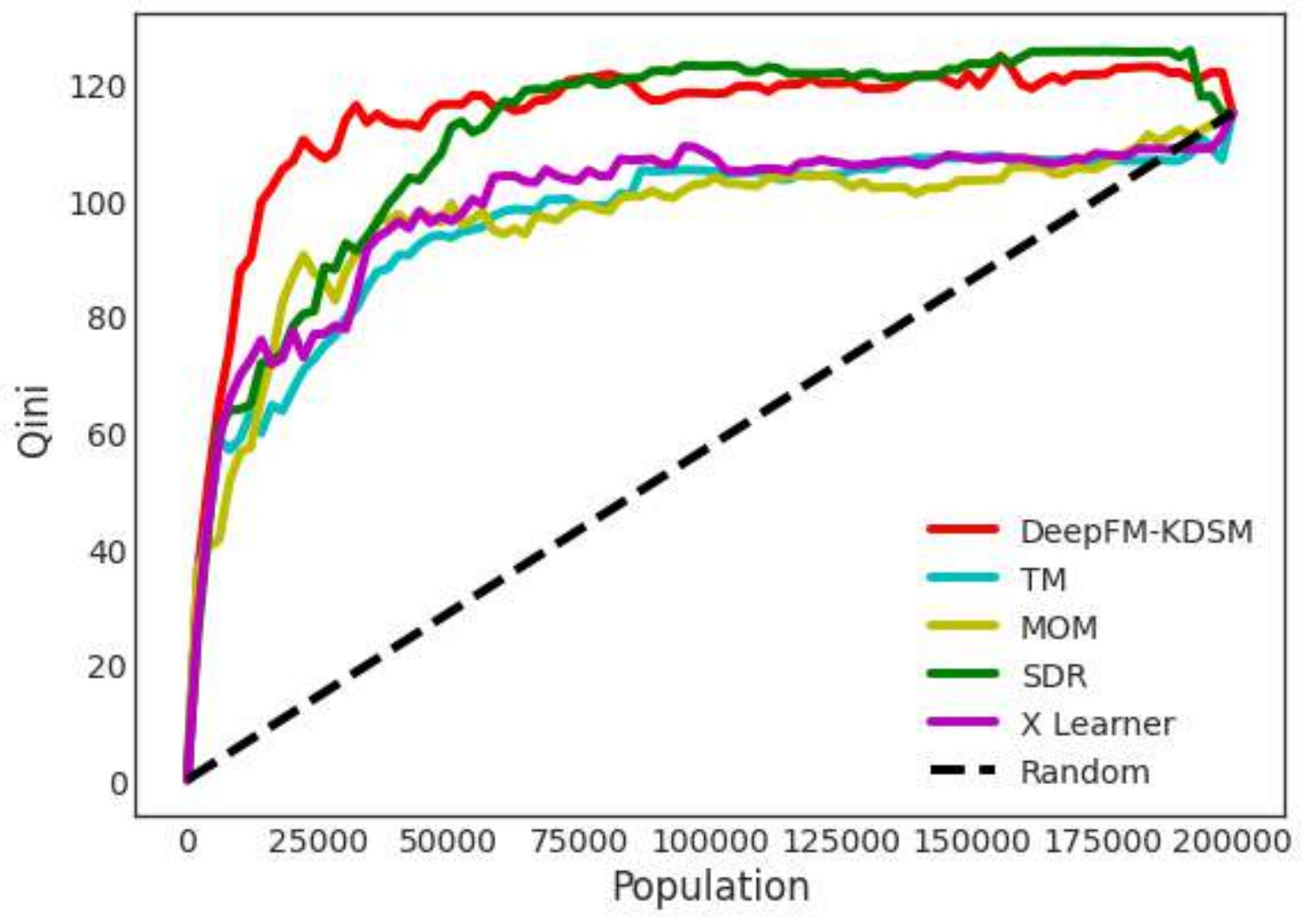}}
    \subfigure[The Qini curves of different methods on the Production dataset.]{
    \includegraphics[width=0.48\linewidth]{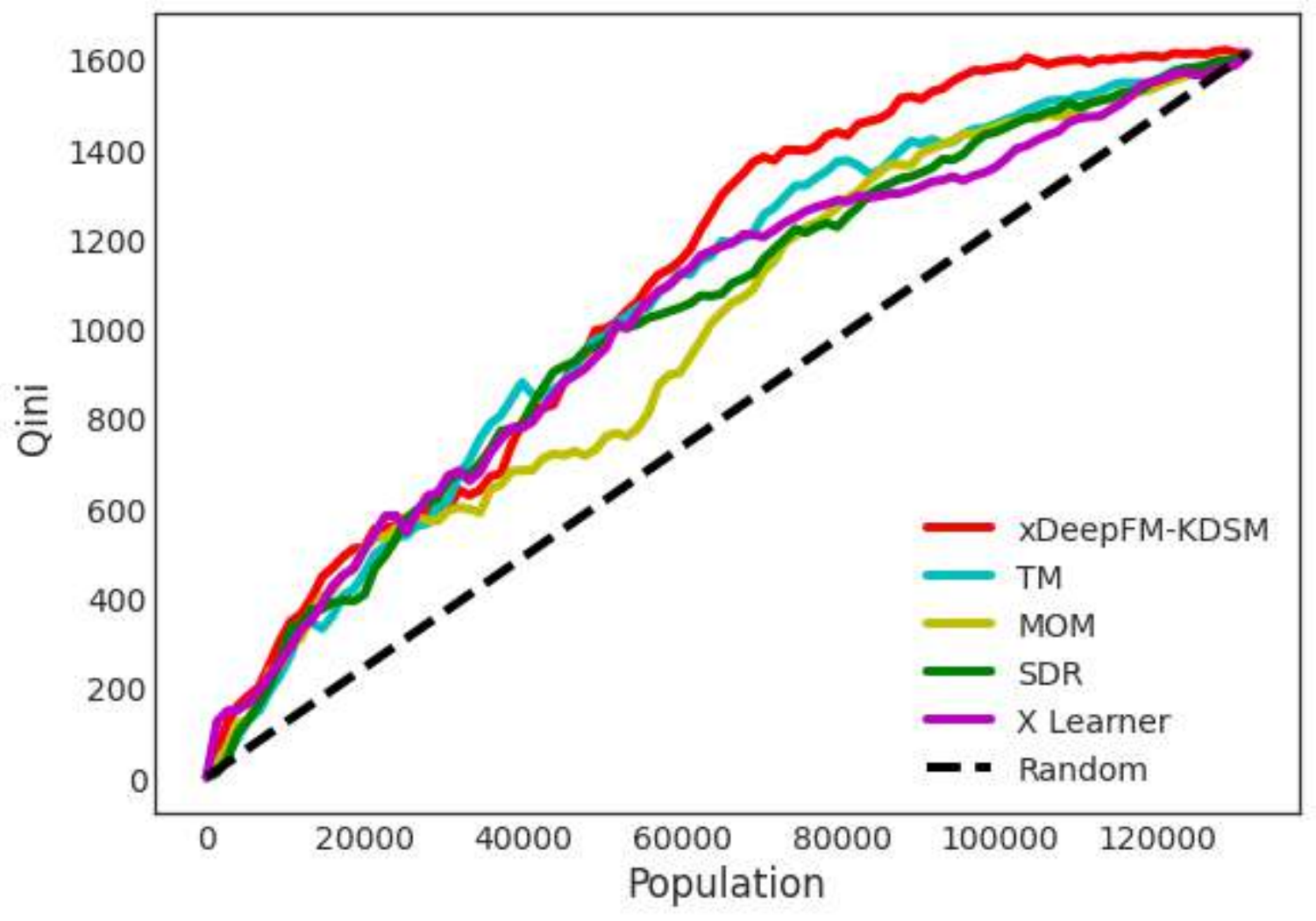}}
    \caption{The Qini curves of different methods.}
    \label{fig5}
\end{figure}

\begin{table}[tbp]\centering
\caption{Performance comparison with different uplift methods.}
\begin{tabular}{l|ll|ll}
\hline
 & \multicolumn{2}{c|}{Criteo} & \multicolumn{2}{c}{Production} \\ \cline{2-5} 
 & \multicolumn{1}{c}{AUUC} & \multicolumn{1}{c|}{Qini} & \multicolumn{1}{c}{AUUC} & \multicolumn{1}{c}{Qini} \\ \hline
TM &  0.8346 &  0.3335 &  0.6470 &  0.1639 \\\hline
MOM & 0.8369 & 0.3363 & 0.6011 & 0.1246 \\\hline
SDR & 0.9616 &  0.4633 &  0.6267 & 0.1459 \\\hline
X-Learner & 0.8565 &  0.3513 &  0.6296 & 0.1397 \\\hline
\textbf{KDSM} & \textbf{0.9784} & \textbf{0.5256} & \textbf{0.6771} & \textbf{0.1890} \\ \hline
\end{tabular}
\label{table5}
\end{table}

As can be observed in Table \ref{table5}, Figure \ref{fig5}, KDSM is capable of targeting persuadable users more precisely. In Figure \ref{fig5}(a), all models can distinguish the top 10,000 users, but DeepFM-KDSM distinguishes the persuadable users who rank about 10,000th-60,000th more accurately than other models. For the population ranking lower than 60,000, DeepFM-KDSM performs similarly to SDR. It is worth mentioning that only DeepFM-KDSM and SDR can distinguish the subgroup ranking about 19,000th-20,000th, to whom the coupon would cause negative effects. In Figure \ref{fig5}(b), xDeepFM-KDSM performs similarly to other methods for the top-ranking users but represents superiority in distinguishing the less persuadable users. In conclusion, KDSM shows its great performance compared with other baselines.

\subsection{Online A/B testing}

We conducted online A/B testing in Qunar Company’s hotel reservation business. The number of room nights is one of the online evaluation metrics of the hotel reservation business, which is the number of reserved rooms × duration (in days). For example, if a user books two rooms for three nights, it would be six room nights. The evaluation metrics of a coupon sending strategy are relative room night increments and cost of coupon for each incremental room night. The target of sending coupons is to bring as many room night increments as possible at a low cost, so a low cost per incremental room night means precise targeting of the persuadable users. In the experiment, the coupon discount is 9\%.

We conducted a randomized controlled experiment with five buckets for seven days on millions of users: (1) Not sending coupons, to get the room night of the control bucket. (2) Sending coupons to all users, to get the maximum increment of room night and the corresponding cost, being the 100\% of other buckets. (3) Sending coupons by Deep Crossing \cite{shan2016deep}. (4) Sending coupons by UpliftDT. (5) Sending coupons by Deep Crossing-KDSM. The relative room night increments and cost of each incremental room night of bucket (2)-(5) are compared with bucket (1). The coupon sending strategy of buckets (3)-(5) is to set an uplift threshold according to the offline Qini curve and send coupons to the users whose uplift prediction exceeds the threshold. The evaluation of online experiments includes the proportion of users who received coupons, the relative room night increments, and the cost per incremental room night of each bucket, as shown in Table \ref{table4}.

\begin{table}[hbp]\centering
\caption{The result of A/B testing. "RN" stands for room night.}
\label{table4}
\begin{tabular}{l|c|c|c}
\hline
\multicolumn{1}{c|}{} & \multicolumn{1}{l|}{\begin{tabular}[c]{@{}l@{}}Proportion \\ of coupons\end{tabular}} & \multicolumn{1}{l|}{\begin{tabular}[c]{@{}l@{}}Relative RN \\ increments\end{tabular}} & \multicolumn{1}{l}{\begin{tabular}[c]{@{}l@{}}Cost per \\ incremental RN\end{tabular}} \\ \hline
(1) No coupons & 0\% & 0\% & 0\% \\ \hline
(2) All users & 100\% & 100\% & 100\% \\ \hline
(3) Deep Crossing & 80\% & 99.50\% & 78.7\% \\ \hline
(4) UpliftDT & 88\% & 99.16\% & 86.0\% \\ \hline
(5) \textbf{KDSM} & 82\% & \textbf{99.70\%} & \textbf{73.5\%} \\ \hline
\end{tabular}
\end{table}

The results show that KDSM can achieve outstanding performance on the room night increments and cost. Compared with sending coupons to all the users (bucket (2)), Deep Crossing-KDSM maintains 99.7\% of the maximum room night increment and costs only 73.5\%. Compared with sending coupons by Deep Crossing w/o KD (bucket (3)), the cost of each incremental room night was reduced by 6.5\% under the KDSM framework.

\subsection{Analysis}

After the extensive offline and online experiments, one may wonder why the performance of KDSM is even better than the teacher model, instead of being between the performance of the teacher and the student. In traditional scenarios of knowledge distillation, researchers usually use a complicated teacher model to distillate a simple student model. Despite their differences in structures, their meaning of output, loss function, and learning goals are the same. 

However, the tasks of the teacher and student model are different but related in the case of KDSM. The teacher UpliftDT maximizes KL or ED to learn the increment, but the student model is optimized by BCE loss and tries to do the binary classification correctly. It can be explained via the idea of multitask learning (MTL) \cite{caruana1997multitask}: MTL aims to improve generalization by leveraging domain-specific information contained in the training signals of related tasks. Thus the teacher model not only does not limit the performance of the student model but also improves its generalization.

Besides, there is a gap between AUUC and the learning objective of the response model. Binary labels of the response model will lead it to learn the probability of purchasing, but what matters to AUUC is the relative order of the uplift value, and simple subtraction would cause error propagation. Thus introducing a soft label of uplift can enhance the sorting precision of the response model.

\section{Conclusion}

In this paper, we study uplift modeling under RCT data and develop a framework called KDSM. We utilize the idea of knowledge distillation and multitask learning to transfer the knowledge of increment into a response model. Additionally, we propose a novel counterfactual sample matching approach in the knowledge distillation framework. Due to the idea of multitask learning, the teacher model is not the upper limit of the student model but provides knowledge from a related task and thus enhances the generalization of the student model. The proposed framework is evaluated on the open-source Criteo dataset and Qunar Company’s real hotel reservation business, and the performance of student models is improved to different extents. In online A/B testing, the cost of each increased room night was reduced by 6.5\% compared with sending coupons by the student model without knowledge distillation. Extensive experiments have proved the superiority of our framework over other baselines. In the future, we may try other sample matching schemes and extend KDSM to multi-treatment scenarios.

\bibliographystyle{unsrt}  
\bibliography{references}

\end{document}